# Spherical Planetary Robot for Rugged Terrain Traversal

Laksh Raura
Space and Terrestrial Robotic Exploration (SpaceTREx) Lab
Arizona State University
781 S Terrace Mall,
Tempe, AZ 85287
laksh.raura@asu.edu

Andrew Warren
Space and Terrestrial Robotic Exploration (SpaceTREx) Lab
Arizona State University
781 S Terrace Mall,
Tempe, AZ 85287
ajwarre3@asu.edu

Jekan Thangavelautham
Space and Terrestrial Robotic Exploration (SpaceTREx) Lab
Arizona State University
781 S Terrace Mall,
Tempe, AZ 85287
jekan@asu.edu

*Abstract*—Wheeled planetary rovers such as the Mars Exploration Rovers (MERs) and Mars Science Laboratory (MSL) have provided unprecedented, detailed images of the Mars surface. However, these rovers are large and are of high-cost as they need to carry sophisticated instruments and science laboratories. We propose the development of low-cost planetary rovers that are the size and shape of cantaloupes and that can be deployed from a larger rover. The rover named SphereX is 2 kg in mass, is spherical, holonomic and contains a hopping mechanism to jump over rugged terrain. A small low-cost rover complements a larger rover, particularly to traverse rugged terrain or roll down a canyon, cliff or crater to obtain images and science data. While it may be a one-way journey for these small robots, they could be used tactically to obtain high-reward science data. The robot is equipped with a pair of stereo cameras to perform visual navigation and has room for a science payload. In this paper, we analyze the design and development of a laboratory prototype. The results show a promising pathway towards development of a field system.

## TABLE OF CONTENTS

## 1. INTRODUCTION

Wheeled planetary rovers such as the Mars Exploration Rovers (MERs) and Mars Science Laboratory (MSL) have provided unprecedented, detailed images of the Mars surface. These rovers have helped to discover sedimentary rock, evidence for past water flow and image rare phenomena unique to Mars, such as the dust devils and "blueberry" mineral chondrites. These rovers provide detailed in-situ images and scientific data. However, these planetary rovers are large, with a mass of 180 kg to 900 kg and are of high-cost. This is required to house sophisticated science instruments and in-situ laboratories. Rapid advancement in miniaturized electronics, power supplies, actuators and even structural materials such as high-strength metallic glass make it possible to develop small, low-cost robots for planetary surface exploration.

In this work, we develop a prototype 2 kg holonomic robot that is spherical and contains a pair of grooved wheels that can traverse over rugged environments (Figure 1). In addition, the robot can also roll unpowered down slopes. The robot has an actuator that enables it to hop 8-50 cm under Martian gravity. The system is powered using a primary or rechargeable battery. With a rechargeable, the total energy in the batteries is 2.3 Wh, while for non-rechargeable 7 Wh can be achieved. The robot communicates wirelessly using radio achieving data rates of up to several Mbps with a nearby ground rover located up to a few kilometers away. Navigation is performed using onboard wheel encoders and a pair of stereo cameras. This is sufficient to perform standard Simultaneous Localization and Mapping (SLAM). Our current prototype utilizes a Raspberry PI, but the platform can use space-qualified radiation hardened computers designed for Mars.

Spherical robots are not new [1-3], however our design combines a hopping mechanism with a holonomic chassis. Thanks to 3D prototyping, we have designed and tested custom grooved wheels to handle rugged terrain. These small, low-cost robots would complement a larger, more capable platform such as the MSL.

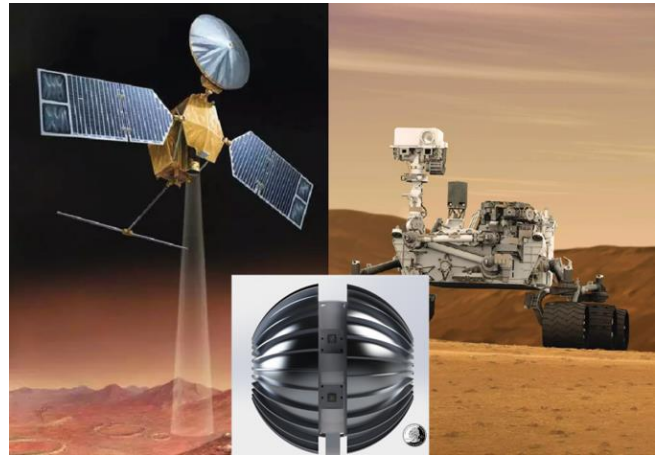

**Figure 1:** Spherical, cantaloupe sized robots (inset) can be deployed from large rovers such as the Mars Science Laboratory and access rugged terrains too risky to traverse by large rovers or be visible to orbital assets such as Mars Reconnaissance Orbiter.

One or more of these robots may be deployed from an MSL-sized rover to explore rugged environments that are inaccessible or too dangerous for a flagship rover. These small robots may be deployed on one-way journeys down slopes, crater rims, canyons and cliffs. Multiple network of robots maybe able to autonomously plan, navigate and explore extreme environments [5, 14, 17-18].

We have developed a laboratory prototype and accompanying low-gravity simulation facility to evaluate the mobility performance of the rover under various Mars and Lunar surface conditions. With these results, we hope to further iterate on the robot design towards a platform ready for extensive testing in the Grand Canyons, Meteor Crater and Flagstaff region of Arizona.

In this paper, we first review past work on small spherical robots for exploration in Section 2, followed by presentation of mission requirements for typical robots of this size in Section 3. In Section 4, we present the robot design followed by brief presentation of the experiment setup in Section 5. This is followed by results and discussion in Section 6 and conclusions in Section 7.

## 2. BACKGROUND

Multiple spherical shaped robots have been proposed for terrestrial and planetary exploration. The RoBall [1] proposed by group from Universite De Sherbrooke is a spherical robot which moves by shifting a suspended mass at its center. This limits the sensors that can be used with the system. Another terrestrial spherical robot, Kickbot [2] was proposed by students at MIT. It uses two external hemispherical shells as wheel for mobility. The concept was designed as a toy. The design has very high maneuverability. Multiple concepts for inflatable spherical robots have been proposed by research teams from Uppsala University, North Carolina State University (NCSU) [3] and University of Toronto. The research team from Uppsala University proposed a design called Spherical Mobile Investigator for Planetary Surface (SMIPS) [4]. These rovers have the advantage of travelling over large distances and steep inclination. However, their reach is limited to benign sandy terrain to avoid damage from sharp rocks.

Other mobility methods were also considered for robot development and it was identified that hopping provides advantage for travelling over longer distances in comparison to rolling alone. This also enable smaller robots to overcome obstacles at least twice the size of the robot. Previous concepts include a series of micro hopping robots developed at MIT [5, 10, 13-14]. These micro-hopping robots would use 'artificial muscle' actuators that are used to energize a spring based hopping mechanism. The robots would be powered using fuel-cells [10, 13, 19] that provides high specific energy. Apart from extreme environment exploration, potential applications for this technology also include terrestrial sensor networks. 'Grillo' is another hopping robot developed at Sant'Anna University [6] and the 7 gram, grass-hopping robot developed at EPFL [7] shows that compact hopping mechanisms can be developed. As an alternative to mechanisms, rocket-powered hopping has also been proposed for planetary exploration [14]. Typically, these mechanisms presented provide very limited control on the direction of the hop. Burdick and Fiorini proposed design for a minimalist jumping robot [8] for planetary exploration. This robot could jump 80 cm and has the ability to control direction of hop. It could leap 40-60 cm based on the angle of projectile. Some other robots like Sandflea [9] by Sandia National Laboratory uses hydraulics for hopping and could hop 50 times its length. But such methods are not viable for application in space or planetary environments. Additionally, the robot could perform maximum 25 hops on a single charge.

## 3. MISSION REQUIREMENTS

The primary goal of this research effort is to develop low-cost, low-mass spherical robots for planetary exploration. The robot would perform the following exploration tasks:

1. Geology by stereo imaging: Identifying size and size distribution of rocks.
2. Wide area investigation: Access wide areas, multiple locations and viewpoints at once using multiple robots to record surface phenomena.

The exploration requirements for the robots are as follow:

1. Ability to traverse over flat, sandy and rocky terrains
2. Access features like pits, craters and cliffs
3. Assist/complement exploration with larger rovers

To accomplish these tasks, the robot require a robust mobility system to travel short distances. To facilitate exploration over a large area, multiple robots may be used. Therefore, each robot would be equipped with a wireless communication system to coordinate exploration and transfer collected data to local server that may be an on-orbit satellite or nearby large rover.

## 4. ROBOT DESIGN

Taking into consideration all the above requirements, our robot design is shown in Figure 2. The inner shell diameter is 15 cm. The inner body is divided into 3 horizontal sections with middle section 4.5 cm thick.

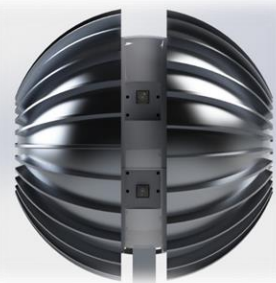

**Figure 2: SphereX Robot showing the wheels, stereo cameras and hopping mechanism.**

The shell thickness was chosen to be 0.3 cm for mechanical strength. The top and bottom section have one camera each, spaced 65 mm apart for stereo imaging. The top section contains the robot control system (Figure 3). The middle section (Figure 4) houses the primary mobility system and the bottom (Figure 5) contains the hopping mechanism.

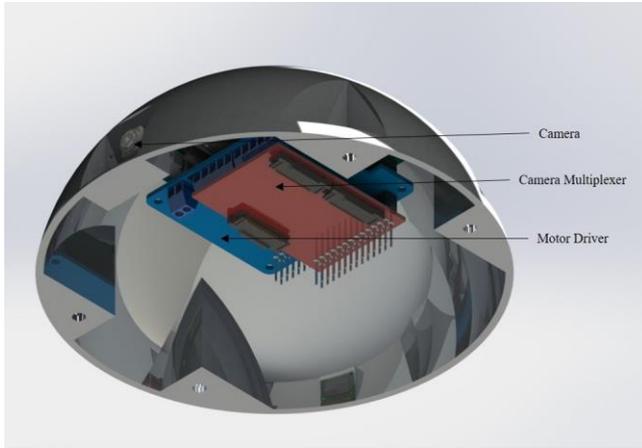

**Figure 3: SphereX Robot top section contains a camera and the control system electronics.**

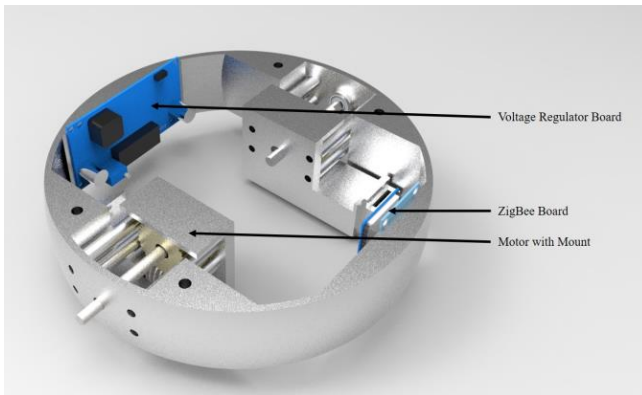

**Figure 4: SphereX Robot middle section contains the motors, communication board and power electronics, with room for a science payload.**

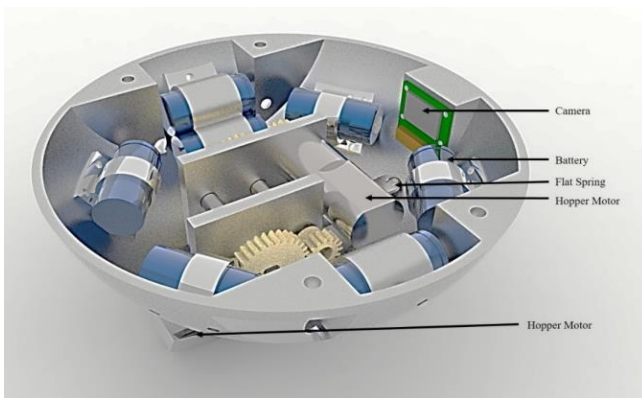

**Figure 5: SphereX Robot bottom section contains the hopping mechanism, batteries and a camera.**

The weight was concentrated in the bottom section which displaced the center of gravity of the robot and ensured mobility by relative motion between wheels and the core section. Figure 6 shows a CAD model of inner shell and exclude the two wheels. Hopping was chosen as the secondary mobility method as it helps to overcome larger obstacles and facilitates faster travel compared to rolling. The two external hemispherical shells (the wheels) were designed with grousers to assist mobility over rocky as well as sandy terrain. These grousers also increase the available traction in low-gravity environments. The wheels are 20 cm in diameter.

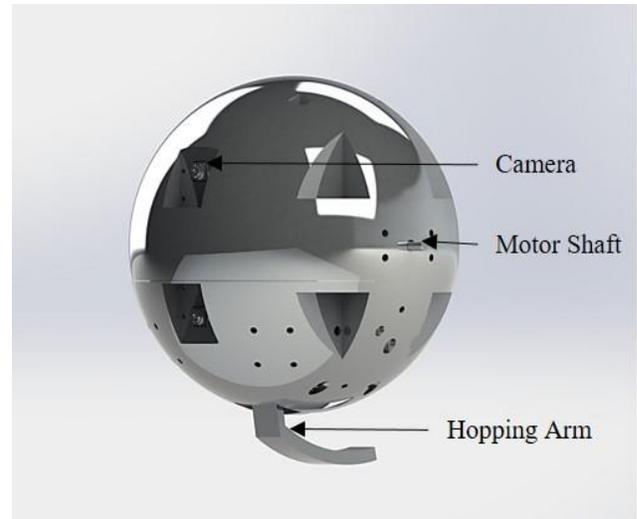

**Figure 6: SphereX body showing the motor shafts, hopping mechanism and stereo camera.**

*External Shell and Drive Train Design*

The drive train was designed for rolling mobility. Table 1 shows the critical parameters for the drive-train design.

**Table 1 - Parameters for Drive Train Design**

| Parameter | | Value |
|---|---|---|
| **g** | Gravity Constant | 1.6 m/s$^2$ |
| **M** | Friction Coefficient (sand) | 0.6 |
| **$\mu_{rr}$** | Rolling Friction (Sand) | 0.15 |
| **$M_r$** | Mass of robot | 2.0 kg |
| **$\theta_s$** | Max grade to be climbed | 14 ° |
| **$V_{max}$** | Maximum linear velocity | 0.03 m/s |
| **$t_a$** | Time to acceleration | 1 s |
| **$R_w$** | Wheel radius | 9.9 cm |
| **$W_n$** | Normal force per wheel | 1.6 N |
| **Rf** | Resistance Factor (grousers) | 20 % |
| **$N_w$** | No. of wheels | 2 |

The motor selection was done on the basis of total traction force required for traversing over a given terrain. The total tractive force is given by Equation 1:

$$F_{TT} = F_{fr} + F_s + F_a \quad (1)$$

Where $F_{TT}$ is the total tractive force, $F_{fr}$ is the frictional force, $F_s$ is the force required to climb the slope and $F_a$ is the force required to accelerate. $F_{fr}$, $F_s$ and $F_a$ are given by Equation (2), (3) and (4) respectively:

$$F_{fr} = M_r * \mu_{rr} \quad (2)$$

$$F_s = M_r * \sin \theta_s \quad (3)$$

$$F_a = M_r * V_{max}/g * T_a \quad (4)$$

Where $M_r$ is the mass of the robot, $\mu_{rr}$ is the rolling friction, $\theta_s$ is the maximum slope to overcome by the robot, $V_{max}$ is the maximum linear velocity and $T_a$ is the maximum time for acceleration. Now, based on the maximum tractive force required, the motor torque is calculated using Equation (5):

$$\tau_r = F_{TT} * D_w/2 * \eta \quad (5)$$

Where $D_w$ is the maximum wheel diameter and $\eta$ is the resistance factor which accounts for the additional friction caused due to grousers on the wheel and free counter rotation of the inner sphere. $\eta$ was chosen to be 20%.

The important factor that influenced selection of motor was maximum linear speed and maximum climb slope. These factors are limited by the maximum traction force the terrain could provide and this can be calculated by

$$F_{T_{max}} = W_n * \mu * D_w \quad (6)$$

Where $F_{T_{max}}$ is the maximum traction force provided by the terrain and $W_n$ is the normal force on each wheel and $\mu$ is the coefficient of friction between the robot wheel surface and terrain.

Based on Equation (5) and (6), a 1000:1 gear ratio is needed and appropriate micro gear motor was selected. Additional reductions were needed to reduce maximum speed of the motor. Therefore, additional reduction of 10:1 was required and thus, a worm gear was added to the system. Figure 4 shows the model of the drive-train for the robot.

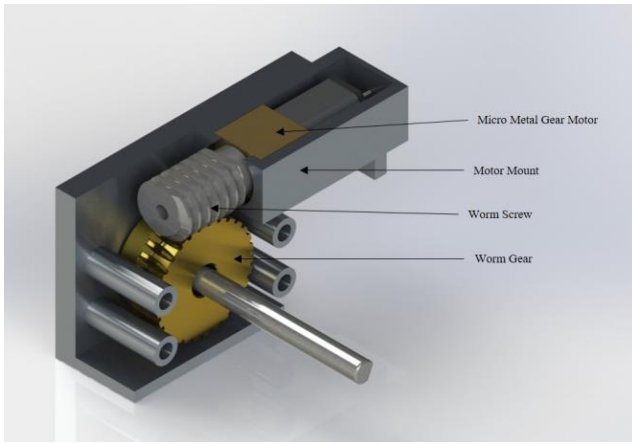

**Figure 6: Motor Drive-train**

The robot wheels were designed to be 20 cm in diameter with the grousers. Grousers help increase the traction on soft soil and help overcome small obstacles. The number of grousers depends on the grouser height and wheel sinkage depth. It is given by the relation defined by Equation (7) [16]:

$$\varphi < \frac{1}{1-i} \left( \sqrt{(1+\hat{h})^2 - (1-\hat{z})^2} - \sqrt{1 - (1-\hat{z})^2} \right) \quad (7)$$

where $\varphi$ is the angle between the two grouser, $i$ is the wheel slip, $\hat{h}$ is normalized height of grouser i.e. ($h/r_w$) and $\hat{z}$ is normalized wheel sinkage i.e. ($z/r_w$). $r_w$ is the radius of the wheel. Based on Equation (7), the angle of separation was calculated in Table 2:

**Table 2: Calculated Separation Angle for Different Grouser Height**

| Grouser Height | $\hat{h}$ | $\hat{z}$ | $\Phi$ |
|---|---|---|---|
| 10 mm | 0.107 | 0.1 | 15.1° |
| 7 mm | 0.074 | 0.08 | 9.4° |

It was assumed only grousers sink in the terrain and thus the maximum height of the grouser was taken as sinkage depth. Using the results from Table 2, the wheels were designed with 24 grousers of 10 mm height and with a separation of 15°.

*Hopping Mechanism Design*

The hopping mechanism enables the robot to overcome an obstacle twice its size. A major challenge was to develop a compact robust system that could be packed inside the robot. Two major requirements for hopping are storage of energy for hopping and mechanism to instantaneously release this energy to perform the hop.

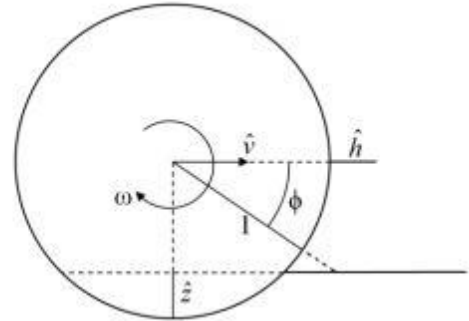

**Figure 7: Diagram to Show the Parameters for Wheel Design.**

Figure 8 shows the CAD model of the designed hopping mechanism. A snail cam was used for charging and instantaneous release of the hopping arm. A cam follower rests on the snail cam and follows the profile of the cam. A follower is connected to the hopping arm. The cam provides maximum displacement of 25 mm or 25.15° radial displacement. The hopping arm is curved to be symmetric with the robot inner sphere. Figure 9 shows a model of the snail cam and hopping arm. The energy is stored in the flat springs. One end of the spring is connected to robot's inner sphere wall and the other rests on the follower such that the hopping arm is loaded.

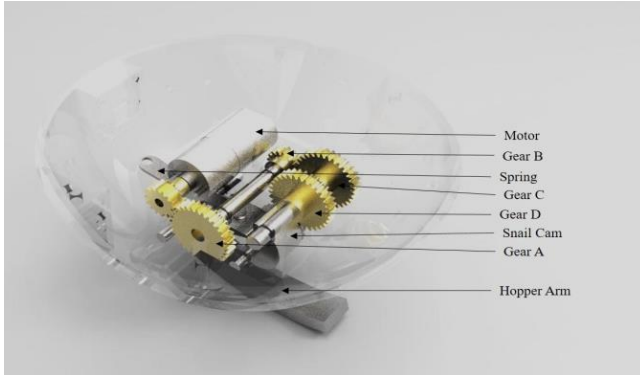

**Figure 8: Model of Designed Hopping Mechanism**

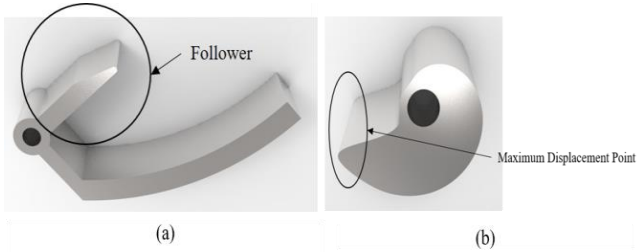

**Figure 9: (a) Model of Hopping arm with cam follower (b) Model of Snail cam**

The mechanism is driven by a geared DC motor and a gear train containing 5 spur gears. The mechanism was designed to provide 20 cm hop on earth. The energy required to hop 20 cm is given by

$$E_{max} = M_r g H_{max} \qquad (8)$$

Where $E_{max}$ is the maximum energy required, $M_r$ is the mass of robot, $g$ is the acceleration due to gravity on earth and $H_{max}$ is the maximum height to be achieved. In an ideal case the, maximum potential energy at maximum height must be equal to energy stored in the spring and can be given by:

$$E_{max} = M_r g H_{max} = \frac{1}{2} k \theta^2 \qquad (9)$$

Where $K$ is the spring constant and $\theta$ is the angular displacement provided by the mechanism. Thus, the required spring constant for the mechanism was calculated to be 71 N/rad. Now, based on calculated spring constant, the counter torque provided by the flat spring would be:

$$\tau_s = k\theta \qquad (10)$$

Where $\tau_s$ is torque applied by the spring. Based on the maximum force:

$$F_s = \tau_s / L \qquad (11)$$

Where $F_s$ is the maximum applied force by spring on cam follower and $L$ is the maximum length of the selected spring. Now to achieve the calculated force total number of spring can be calculated:

$$n = \Psi \, FL^3 / Esbt^3 \qquad (12)$$

Where,

$$\Psi = 3 / (2 + \frac{\acute{n}}{n}) \qquad (13)$$

Where $E$ is the Young's modulus of spring, $L$ is length of spring, $s$ is maximum deflection $b$ is maximum width of spring and $t$ is the thickness of spring. $n´$ is number of spring of equal length. We have taken all the spring of equal length. From these calculations, 6 springs are required. Figure 10(a)-10(d) shows the operation of the hopping mechanism.

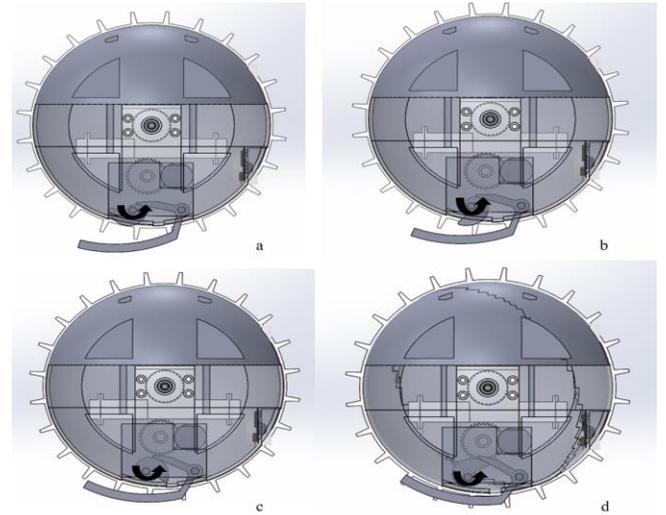

**Figure 10: Operation of Hopping Mechanism (a) Maximum extension of arm for hopping (b)-(d) Rewinding of the hopping mechanism for next hop**

Table 3 shows the robot mass budget. The heaviest components include the chassis and the hopping mechanism. As this is the first iteration of the hopping mechanism, we hope to further reduce its mass during future iterations. Other components including the electronics and motors take up very little mass. Table 4 shows the nominal power budget for SphereX.

**Table 3: Mass Budget for SphereX Prototype**

| Subsystem | Unit | Margin | Total Mass (g) |
|---|---|---|---|
| Structure | System Chassis | 1.4 | 300 |
| Onboard Computer | Raspberry Pi Board | 1.1 | 25 |
| Peripheral Microcontroller | Arduino | 1.1 | 10 |
| Communications | Zigbee Board | 1.1 | 10 |
| Primary Mobility System | Motors | 1.3 | 105 |
| | Control Board | 1.1 | 25 |
| | Wheels | 1.3 | 35 |
| Second Mobility System | Hopping Mechanism | 1.3 | 540 |
| | Springs | 1.2 | 15 |
| Sensors | Cameras | 1.1 | 10 |
| | Camera Multiplexer | 1.1 | 15 |
| Power System | Batteries | 1.2 | 165 |
| | Power Regulator | 1.1 | 45 |
| Total Mass | | | 1300 g |
| Mass Limit | | | 2000 g |
| Mass Margin | | | 35 % |

**Table 4: Power Budget for Designed Robot**

| Unit | Instrument Duty Cycle | Power Calculated (W) | Allotted Power (W) | Margin | Total Energy Required ($10^{-1}$ Wh) |
|---|---|---|---|---|---|
| Electronics | 1 | 7.15 | 8.50 | 18.88 | 7.15 |
| Motor | 1 | 11.54 | 13.50 | 16.94 | 11.54 |
| Camera | 1 | 2.08 | 2.50 | 20.19 | 2.08 |
| Radio | 1 | 0.18 | 0.22 | 19.05 | 0.18 |
| Hopping | 0.2 | 5.61 | 6.50 | 15.74 | 1.12 |
| Total Energy Consumed Per Hour | | | | | 2.2 Wh |
| Total Energy Available from Battery | | | | | 1.9 Wh |
| Operation Time | | | | | 52 min |

## 5. EXPERIMENT SETUP

To test the performance of the robot mobility system in a low gravity environment, a low-gravity simulation testbed needed to be designed. Previously, robot suspension systems have been developed to simulate operations in a low-gravity environment. Active Response Gravity Offload System (ARGOS) [11] developed by NASA is one such system. Also in the past, mobile suspension systems have been proposed and used for simulating low gravity environment outdoors [12]. We designed a Low-gravity Offset and Motion Assistance and Simulation System (LOMASS) shown in Figure 11.

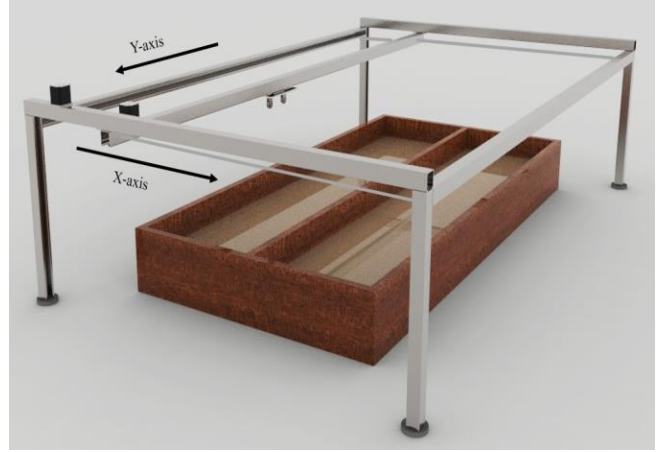

**Figure 11: Low-gravity Offset and Motion Assistance and Simulation System (LOMASS).**

The system contains a 2 m long and 1 m wide box which was used to simulate different terrains and slopes. The setup contains an overhead automated gantry which suspends the robot in the box. Both axes are driven by belt drive and stepper motors. The gantry is controlled by an Arduino and two stepper drivers. The speed of the overhead gantry was matched with the set speed of the robot in an open loop and manual setting. This was to make sure the gantry stays over the robot. A closed loop system is being developed for future testing of robot. The Table 5 shows the specification of the LOMASS system.

**Table 5: LOMASS Specifications**

| Dimensions | 2.4 m x 1.2 m x 0.6 m |
|---|---|
| Max Travel Distance X – axis | 0.75 m |
| Max Travel Distance Y – axis | 1.80 m |
| Max traverse speed X –axis | 10 m/min |
| Max traverse speed Y-axis | 20 m/min |

For our experiments, a representative 3D printed robot was made. Two sets of wheel were printed with 7 mm and 10 mm grouser height. For testing two sets of terrain were created in LOMASS. The SphereX robot was tested in loose sand as well as graveled and rocky terrain. Each robot was also tested at a slope of 10°. Each run was 1.4 m long. The time and power required for travel was measured for each run. A hopping mechanism was integrated into the robot prototype. Hopping was tested on a hard surface and height of hop was measured. The robot was suspended using pulley and offset mass to simulate performance of hopping mechanism under simulated Martian gravity conditions.

## 6. RESULTS AND DISCUSSION

The velocity of the robot was measured using encoders mounted on wheel shafts. For the experiments, the robot was made to follow a straight line path. The time for travelling 1.4 m length was calculated and used to determine the average velocity. The power consumption was measured using an in-loop INA219 current sensor. The aim of the experiment was to show that robot would be able to move and hop in a low gravity environment.

*Performance under Simulated Lunar Gravity*

The robot was first tested on a level sandy terrain. Figure 10 shows the plot of velocity and power consumption of the robot over the total run time. The average standby power was 15 W and ~20 W was consumed while in motion. The set speed for robot was 1.5 m/min. The wheel speed was maintained at the set speed using a PI controller. The robot took 75 seconds to travel 1.4 m, thus, the average speed of travel was calculated to be 1.1 m/min. Therefore, the average slip on level surface under lunar gravity was 23 %. Figure 11 shows some points with low speed and higher power and vice versa. This may be due to certain uneven patches in the travel path.

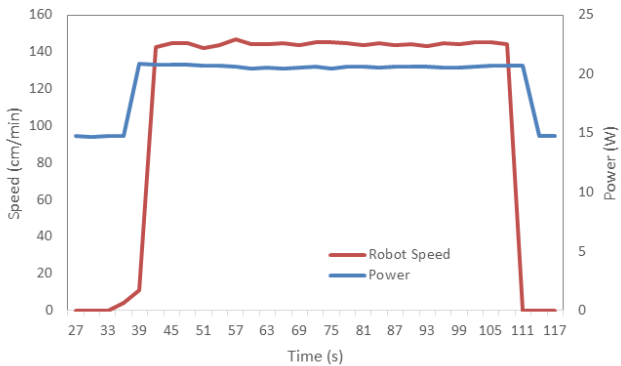

**Figure 10: Robot velocity and power vs time for levelled sandy surface under lunar gravity.**

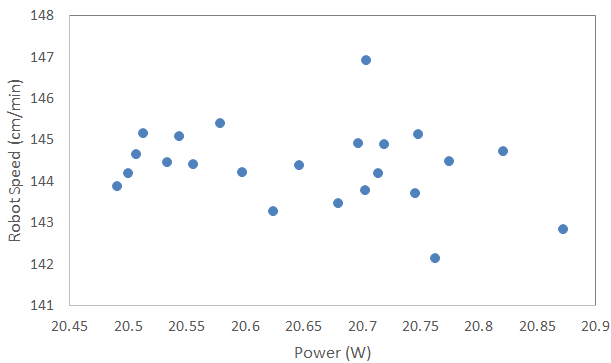

**Figure 11: Robot velocity vs power for levelled sandy surface, 10 mm Grouser wheels under lunar gravity.**

The plot of robot velocity vs. power shows the how power consumption varies with respect to velocity achieved and also provides details regarding approximate variation in terrain. A 10 ° slope was created on sandy surface and performance was evaluated. Figure 12 shows the performance on a sloped surface.

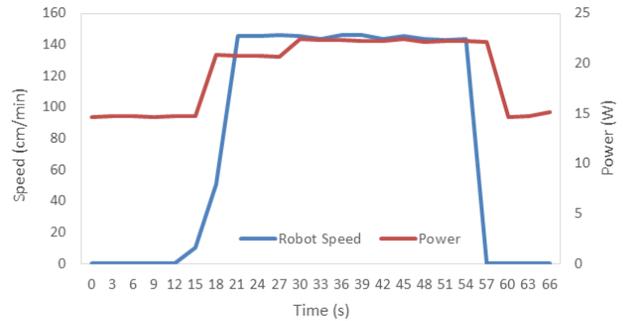

**Figure 12: Robot velocity and power consumption vs time for 10 ° slope under lunar gravity.**

The total length of run was 50 cm with 40 cm long slope. The total traversal time was 31 seconds and thus, the average speed was 0.77 m/min. The approximate slip was 47 %. The average power required on the slope was 22.3 W (Figure 13).

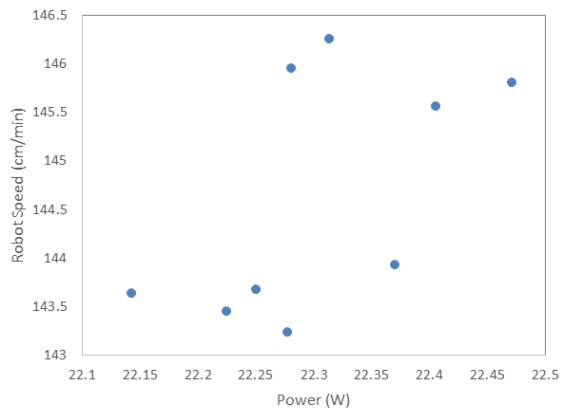

**Figure 13: Robot velocity vs power for 10 ° slope on sandy surface, using 10 mm grouser wheels under lunar gravity.**

Figure 14 shows the performance of the robot on rocky and graveled terrain. The robot required 80 seconds to travel 1.4 m. The average speed of robot was 1.05 m/min and the resultant slip was, thus, 29 %. It was slightly higher compared to sandy terrain. There was a variation in slip percentage of approximately 5% over multiple runs. This was because of discontinuous traction. The average power required was ~21.4 W (Figure 15).

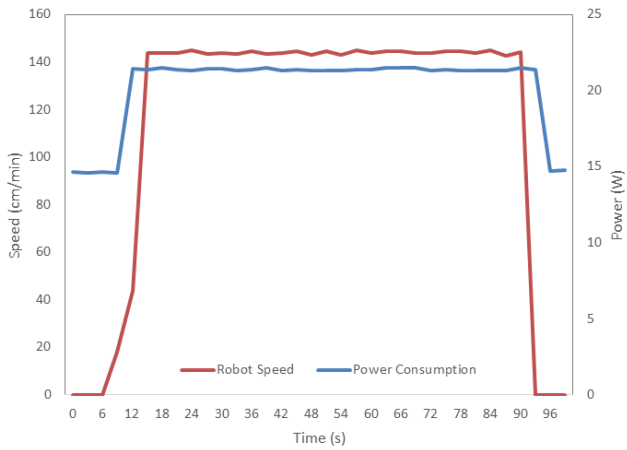

**Figure 14: Robot velocity and power consumption vs time for rocky surface under lunar gravity.**

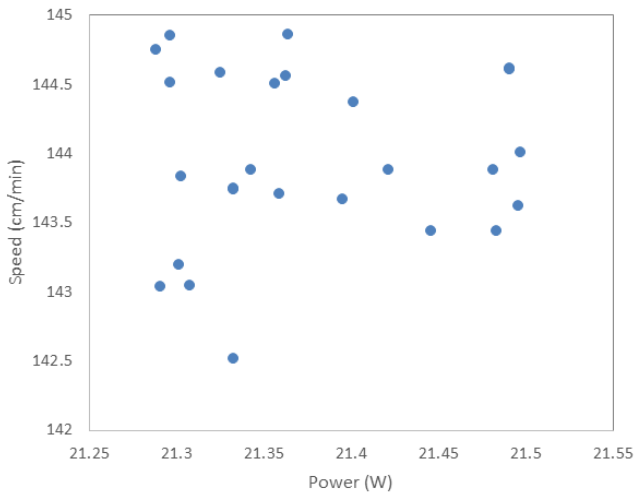

**Figure 15: Robot velocity vs power for rocky surface, 10 mm grouser wheels under lunar gravity.**

Now, with 7 mm high grouser wheels, the robot performance was evaluated on a levelled sandy surface. There was an increase in the average speed and thus, reduction in average slip. The slip was approximately 15%. This may be due to higher number of grousers and smaller angle of separation between them. The power consumption for the experiment was 20.6 W.

*Performance under Simulated Martian Gravity*

With 10 mm high grouser wheels and levelled sandy surface, the experiment was repeated under simulated Martian gravity. The average velocity of robot was 1.33 m/min. The average slip was calculated to be approximately 7% which is significantly lower compared to lunar gravity. There was a small increase in the average power and it was found to be 21.9 W.

*Performance at Hopping System*

Figure 16 shows one of our preliminary hopping tests. It was observed that robot could hop 8 – 16 cm under simulated Martian gravity. The average power consumption for each hop was 16 W and each cycle was 3 second. However our preliminary design utilized 3D printed plastic and it was damaged after several hops. Plans are underway to machine an Aluminum Al-6065 chassis and that should overcome the problems faced with the 3D printed prototype. With the metal chassis, we should be able to add more spring to further increase the hopping height to the desired 25 to 50 cm.

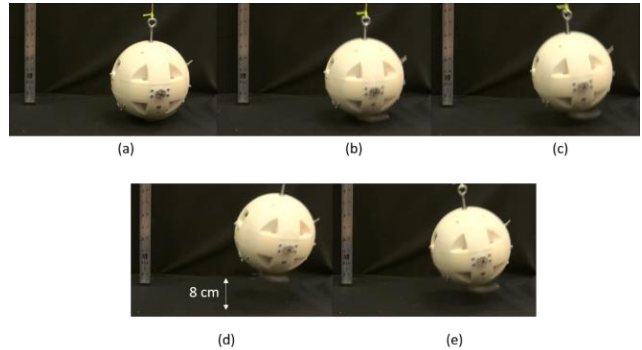

**Figure 16: Test for Operation of Hopping Mechanism at Simulated Martian Gravity**

## 7. CONCLUSIONS

A new spherical micro robot called SphereX has been proposed. A fully working prototype has been designed and built. The prototype has been tested under simulated lunar and Martian gravity conditions. The robot was also tested under sandy and rugged terrain. The robot mobility performance was found to be good. It was observed that as angle of separation between grouser decreases there is increase in average speed of robot and the power consumption remains almost constant. A hopping mechanism was developed for the robot that enables the robot to in theory perform unlimited hops. Currently the system is able to perform a hop of 8 - 10 cm under simulated Martian gravity. Extrapolating this, we would be able to achieve 16 - 20 cm hop in lunar conditions. The performance of hopping mechanism has to be improved to achieve the stated mission requirements. Based on power consumption for each hop and maximum power available, it was calculated that the robot would be able to produce maximum 208 hops in a single charge and robot would operate for 35 minutes of continuous hopping. The proposed SphereX design shows a promising pathway towards further maturation and testing of the technology in the field.

## BIOGRAPHY

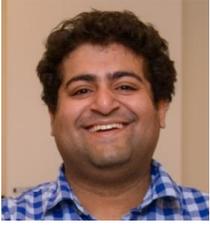
***Laksh Raura*** *received a M.S. in Engineering from Arizona State University, Tempe in 2016. He is currently working with the Space and Terrestrial Robotic Exploration (SpaceTREx) Laboratory at Arizona State University as a Mechatronics System Engineer. He is involved as an instrument engineer on the Asteroid Origins Satellite - 1 (AOSAT-1), a CubeSat mission set for launch in mid-2017. He started his career as an Application Engineer at B&R Industrial Automation and moved to ASU to pursue his Master. His areas of interest include extreme environment robotics, mechatronic system design and mechanism for space application.*

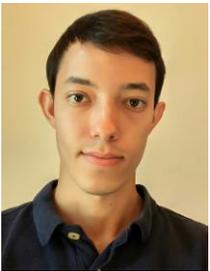
***Andrew Warren*** *is a senior undergraduate student in Electrical Engineering at Arizona State University. The focuses of his studies are in solid state electronics and electrical material properties, which he intends to continue in pursuit of a Masters. He has been involved in research since his freshman year, working on developments in fields such as extreme environment robotics and space systems. Notable projects he has contributed to include the Micro Submersible Lake Exploration Device (MSLED), the Goddard Remotely Operated Vehicle for Exploration and Research (GROVER), the Asteroid Origins Satellite I (AOSat I), and the Space Weather and Impact Monitoring Satellite (SWIMSat).*

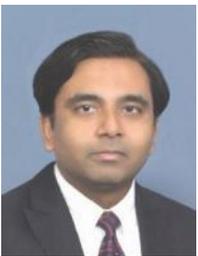
***Jekan Thangavelautham*** *is an Assistant Professor and has a background in aerospace engineering from the University of Toronto. He worked on Canadarm, Canadarm 2 and the DARPA Orbital Express missions at MDA Space Missions. Jekan obtained his Ph.D. in space robotics at the University of Toronto Institute for Aerospace Studies (UTIAS) and did his postdoctoral training at MIT's Field and Space Robotics Laboratory (FSRL). Jekan Thanga heads the Space and Terrestrial Robotic Exploration (SpaceTREx) Laboratory at Arizona State University. He is the Engineering Principal Investigator on AOSAT I and is a Co-Investigator on SWIMSat, a CubeSat mission to monitor space threats.*